\documentclass[twoside,12pt]{article}
\usepackage[hyphens]{url} 
\usepackage[colorlinks=true,linkcolor=blue,citecolor=blue]{hyperref}
\usepackage[%
    style=numeric-comp,sorting=none,
    sortcites=true,doi=false,url=false,
    giveninits=true,maxbibnames=99,hyperref]{biblatex}
\usepackage{fancyhdr}
\usepackage[headheight=18pt,margin=1in]{geometry}
\usepackage[utf8]{inputenc}
\usepackage{framed,multirow}

\usepackage{amssymb}
\usepackage{amsmath}
\usepackage{graphicx}
\usepackage{enumitem}
\usepackage{subfigure}
\usepackage{booktabs} 
\usepackage{caption}

\usepackage{titlesec}
\titlespacing*{\section}{0pt}{1.1\baselineskip}{\baselineskip}

\title{\vspace{-1cm}Image Classification using Graph Neural Network\\ and Multiscale Wavelet Superpixels}
\date{}

\newcommand{\red}[1]{\textcolor{red}{#1}}
\newcommand{\bx}{\boldsymbol{x}}
\newcommand{\bxs}{\boldsymbol{x_s}}
\newcommand{\bxS}{\boldsymbol{x_S}}

\newcommand{\bxo}{\boldsymbol{x_0}}
\newcommand{\xsb}{\boldsymbol{x_s}}

\usepackage{scalerel,stackengine}
\stackMath
\newcommand\widecheck[1]{%
\savestack{\tmpbox}{\stretchto{%
  \scaleto{%
    \scalerel*[\widthof{\ensuremath{#1}}]{\kern-.6pt\bigwedge\kern-.6pt}%
    {\rule[-\textheight/2]{1ex}{\textheight}}
  }{\textheight}%
}{0.5ex}}%
\stackon[1pt]{#1}{\scalebox{-1}{\tmpbox}}%
}

\newcommand\widecheckup[1]{%
\savestack{\tmpbox}{\stretchto{%
  \scaleto{%
    \scalerel*[\widthof{\ensuremath{#1}}]{\kern-.6pt\bigwedge\kern-.6pt}%
    {\rule[-\textheight/2]{1ex}{\textheight}}
  }{\textheight}%
}{0.5ex}}%
\stackon[1pt]{#1}{\scalebox{1}{\tmpbox}}%
}
\usepackage{url}
\usepackage{xcolor}
\definecolor{newcolor}{rgb}{.8,.349,.1}

\begin{document}

\maketitle

\vspace{-3cm}
\begin{center}
Varun Vasudevan$^{a,}$\footnote{\label{note1}Equal contribution}, Maxime Bassenne$^{b,}$\textsuperscript{\ref{note1}}, Md Tauhidul Islam$^{b,*}$, and Lei Xing$^b$

$^a$Institute for Computational and Mathematical Engineering, Stanford University, Stanford, CA-94305, USA

$^b$Department of Radiation Oncology, Stanford University, Stanford, CA-94305, USA

*Corresponding author. Email: tauhid@stanford.edu
\end{center}

\section*{Abstract}
Prior studies using graph neural networks (GNNs) for image classification have focused on graphs generated from a regular grid of pixels or similar-sized superpixels. In the latter, a single target number of superpixels is defined for an entire dataset irrespective of differences across images and their intrinsic multiscale structure. On the contrary, this study investigates image classification using graphs generated from an image-specific number of multiscale superpixels. We propose WaveMesh, a new wavelet-based superpixeling algorithm, where the number and sizes of superpixels in an image are systematically computed based on its content. WaveMesh superpixel graphs are structurally different from similar-sized superpixel graphs. We use SplineCNN, a state-of-the-art network for image graph classification, to compare WaveMesh and similar-sized superpixels. Using SplineCNN, we perform extensive experiments on three benchmark datasets under three local-pooling settings: 1) no pooling, 2) GraclusPool, and 3) WavePool, a novel spatially heterogeneous pooling scheme tailored to WaveMesh superpixels. Our experiments demonstrate that SplineCNN learns from multiscale WaveMesh superpixels on-par with similar-sized superpixels. In all WaveMesh experiments, GraclusPool performs poorer than no pooling / WavePool, indicating that poor choice of pooling can result in inferior performance while learning from multiscale superpixels.


\begin{figure*}[!htb]
    \centering	
    \includegraphics[clip,angle=90,scale=0.6]{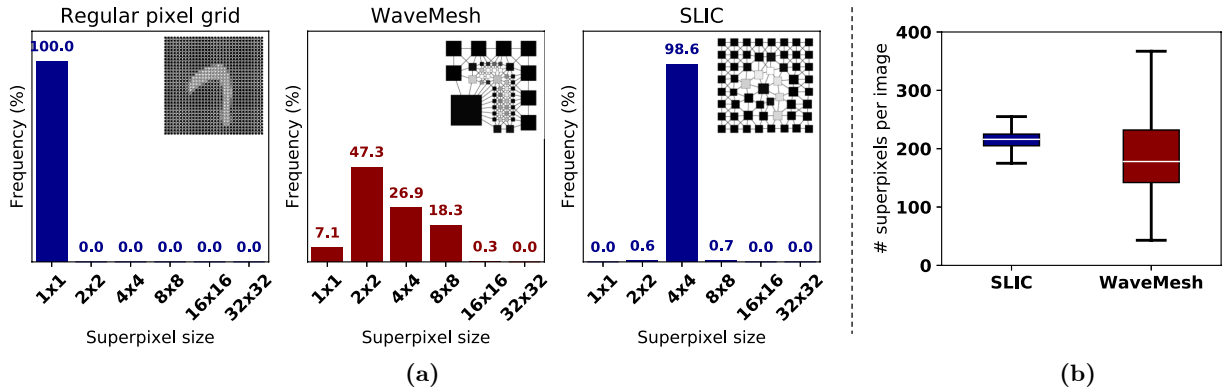}
    \caption{(a) Average distribution of superpixel size averaged across MNIST training dataset for different superpixel representation: none (left), WaveMesh (center), and similar-sized SLIC superpixels (right). In each panel, an insert shows the graph representation of a single sample for illustration. Size of a node in the graph is proportional to the superpixel size. SLIC superpixels are not cubic yet the x-axis binning is chosen to match other plots. (b) Boxplots of the \# superpixels per image for CIFAR-10 training dataset.}\label{fig_overview}
\end{figure*}

\section{Introduction}
\label{intro}

Convolutional neural networks (CNNs) achieve the best performance on various image classification tasks. CNNs learn to classify images from a regular pixel-grid representation of the image. Two limitations of this approach are: 1) Although not all pixels provide an equal amount of new information, by design, the filters in the first layer of a CNN operate on each pixel from top-left to bottom-right in the same way; 2) CNNs require images to be of the same size. Therefore, images are typically resized to a prescribed size before feeding into a CNN. In applications that use standard CNN architectures or pre-trained models on a new image classification dataset, the images are typically uniformly downsampled to meet the input size requirements of the architecture being used. Uniform downsampling may be suboptimal as real data naturally exhibits spatial and multiscale heterogeneity. Few studies have explored the impact of input image resolution on model performance \cite{sabottke2020effect}, despite its recognized importance \cite{lakhani2020importance}.

In contrast to CNNs, graph neural networks (GNN) learn from a graph representation of the image. Several studies have shown promise in classifying images from graphs using GNNs \cite{monti2017geometric, fey2018splinecnn, knyazev2019image, knyazev2019understanding,  dwivedi2020benchmarking,avelar2020superpixel,mesquita2020rethinking}. Unlike CNNs, the GNNs used in these studies do not require input graphs to have the same size/structure (e.g., number of nodes and edges can be different). However, these studies have been restricted to graphs that represent either a regular grid of pixels or similar-sized superpixels. In the latter, a single target number of superpixels is defined for an entire dataset irrespective of differences across images and their intrinsic multiscale structure. 

In summary, while GNNs do not impose any restrictions on the size of superpixels in an image, prior studies have not systematically explored classifying images from graphs that represent multiscale superpixels. In this paper, we fill this gap by investigating image classification using a multiscale superpixel representation that can be considered as in between the regular-grid and similar-sized superpixel representations as shown in \autoref{fig_overview}. Our contributions are as follows.
\begin{itemize}[itemsep=-2pt,topsep=0pt,leftmargin=*]
\item We present WaveMesh, an algorithm to superpixel (compress) images. WaveMesh is based on the quadtree representation of the wavelet transform. Our sample-specific method leads to non-uniformly distributed multiscale superpixels. The algorithm systematically computes the number and size of superpixels in an image based on the image content. WaveMesh requires at most one tunable parameter. WaveMesh superpixels allow us to rethink the process of downsampling (superpixeling) images to a fixed size. 
\item We propose WavePool, a spatially heterogeneous pooling method tailored to WaveMesh superpixels. WavePool preserves spatial structure leading to visually interpretable intermediate graphs. WavePool generalizes the classical pooling employed in CNNs and easily integrates with existing GNNs.
\item We compare the performance of WaveMesh and similar-sized superpixels using SplineCNN, a state-of-the-art network for image graph classification \cite{fey2018splinecnn}. Using SplineCNN, we perform extensive experiments on three benchmark datasets under three local-pooling settings: no pooling, GraclusPool, and WavePool. 
\end{itemize}

\section{Related work}\label{relatedwork}

\textbf{Superpixeling.} Grouping pixels to form superpixels was proposed as a preprocessing mechanism that preserves most of the structure necessary for image segmentation \cite{ren2003learning}. For a detailed review and evaluation of various state-of-the-art superpixeling algorithms see \cite{stutz2018superpixels} and \cite{giraud2019evaluation}. Few of them are ERS, SLIC, SEEDS, MSS, ERGC, LSC, ETPS, and SCALP. These algorithms generate similar-sized superpixels, and were originally developed and evaluated in the context of image segmentation and not image classification.

\textbf{GNN for image graph classification.} Many studies have demonstrated the representational power and generalization ability of GNNs on image graph classification tasks using similar-sized superpixels. SplineCNN is a network for learning from irregularly structured data that builds on the work of MoNet \cite{monti2017geometric}, but uses a spline convolution kernel instead of Gaussian mixture model kernels \cite{fey2018splinecnn}. Recognizing the importance of spatial and hierarchical structure inherent in images, Knyazev et al.\ model images as multigraphs that represent similar-sized superpixels computed at different user-defined scales, and then successfully train GNNs on the multigraphs \cite{knyazev2019image}. Dwivedi et al.\ show that message passing graph convolution networks (GCN) outperform Weisfeiler-Lehman GNNs on MNIST and CIFAR-10 datasets \cite{dwivedi2020benchmarking}. 

\textbf{Local pooling.} Local pooling is used in GNNs to coarsen the graph by aggregating nodes within specified clusters \cite{mesquita2020rethinking}. Graclus is a kernel-based multilevel graph clustering algorithm that efficiently clusters nodes in large graphs without any eigenvector computation \cite{dhillon2007weighted}. Graclus is used in many GNNs to obtain a clustering on the nodes, which the pooling operator then uses to coarsen the graph \cite{defferrard2016convolutional, monti2017geometric, fey2018splinecnn,mesquita2020rethinking}. Hereafter, we refer to pooling based on Graclus clustering as GraclusPool. Mesquita et al. show that convolutions play a leading role in the success of GNNs and not local pooling \cite{mesquita2020rethinking}. 

\begin{figure}[ht!]
    \centering	
    \includegraphics[width=\textwidth]{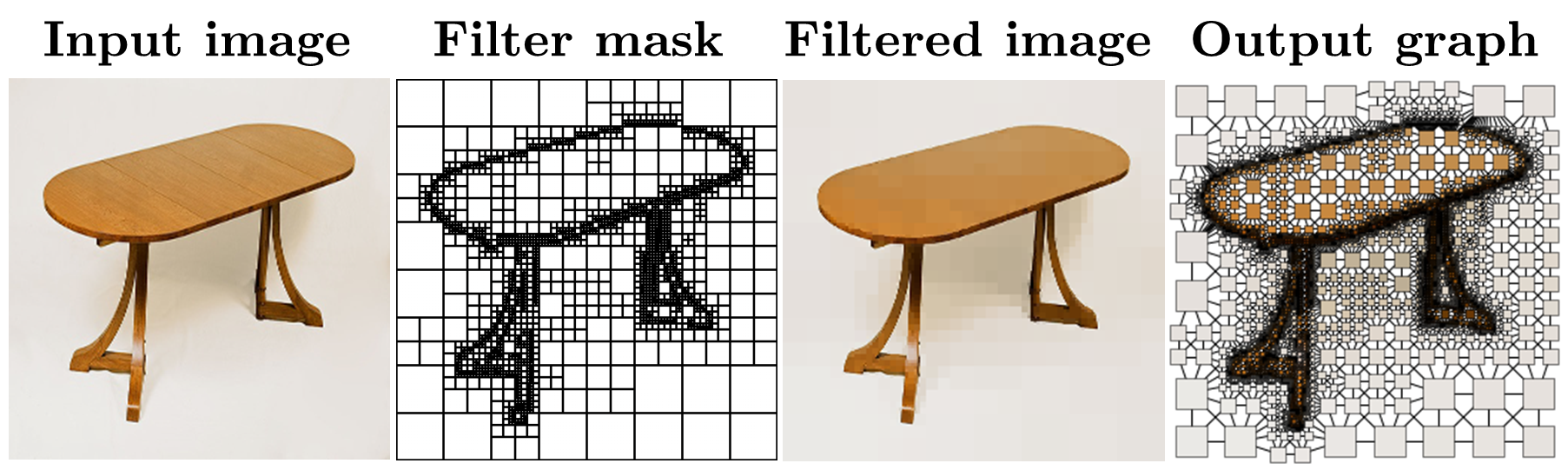}
    \caption{Filtering image in wavelet space generates a non-uniform superpixel mesh which is then represented as a graph. Input image is processed with the method described in \autoref{proposedmethod} with a threshold equal to five times the theoretical value.}\label{fig_framework}
\end{figure}

\section{WaveMesh: Multiscale Wavelet Superpixel}\label{prop_sp}
\label{proposedmethod}

The WaveMesh algorithm is broken down into its elementary steps below: 1) images are wavelet transformed, 2) images are filtered in wavelet space by thresholding the wavelet coefficients, and 3) the superpixel mesh is generated from the wavelet-filter mask. The algorithm is rooted in wavelet theory's seminal work \cite{Mallat89,donoho1994ideal}. The particular way in which wavelets are used in this work is inspired by their related application in the physical sciences \cite{Schneider10,Bassenne18wave}.

\subsection{Step 1: Wavelet Transform of the Input Image}
\label{prop_sp_WT}

Consider a two-dimensional (2D) image $I$ discretely described by its pixel values $I[\bxo]$ centered at locations $\bxo=2^{-1}(i\Delta,j\Delta)$ of a $N{\times}N$ regular grid, where $\Delta$ is the inter-pixel spacing and $(i,j)= 1, 3, \dots, 2N-1$. A continuous wavelet representation of $I$ is $I(\bx)=\sum_{\bxo}\widecheckup{I}^{(0)}[\bxo]\phi^{(0)}(\bx-\bxo)$, where $\bx$ is the continuous pixel-space coordinate, and $\phi^{0}(\bx-\bxo)$ are scaling functions that form an orthonormal basis of low-pass filters centered at $\bxo$, with filter width $\Delta$. The scaling functions have unit energy $\left\langle \phi^{0}(\bx-\bxo) \phi^{0}(\bx-\bxo)\right\rangle=1$, where the bracket operator $\langle y \rangle = 1/(N \Delta)^2 \int y(\bx) d\bx$ denotes the global average for a general 2D continuous field $y(\bx)$. In practice, when dealing with discrete signals, $\widecheckup{I}^{(0)}[\bxo]$ cannot be computed exactly, since $I$ is only known at discrete points $\bxo$. Instead, it is numerically discretized and the approximation coefficients $\widecheckup{I}^{(0)}[\bxo]$ are estimated as an algebraic function of $I[\bxo]$. Assuming that $\phi^{0}(\bx-\bxo)$ decays fast away from $\bx=\bxo$, we get 
$\widecheckup{I}^{(0)}[\bxo]=I[\bxo]/N$ \cite{Adison}. This estimate for $\widecheckup{I}^{(0)}[\bxo]$  is the initialization stage of the recursive wavelet multiresolution algorithm (MRA) \cite{Mallat89}, which enables the computation of wavelet coefficients at coarser scales.

The decomposition of the finest-scale low-pass filter $\phi^{0}(\bx-\bxo)$ in terms of narrow-band wavelet filters $\psi^{(s,d)}(\bx-\bxs)$ with increasingly large filter width and a coarsest-scale scaling function $\phi^{(S)}(\bx-\bxS)$ yields the full wavelet-series expansion of $I$,
\begin{equation}
\vspace{-3pt}
I(\bx)=\sum_{s=1}^{S}\sum_{\bxs}\sum_{d=1}^{3}\widecheck{I}^{(s,d)}[\bxs]\psi^{(s,d)}(\bx-\bxs)+\widecheckup{I}^{(S)}[\bxS]\phi^{(S)}(\bx-\bxS). \label{W4}
\end{equation}
Here, $\widecheck{I}^{(s,d)}[\bxs]=\left\langle I(\bx)\psi^{(s, d)}(\bx-\bxs)\right\rangle$ and $\widecheckup{I}^{(S)}[\bxS]=\left\langle I(\bx)\phi^{(S)}(\bx-\bxS)\right\rangle$
are wavelet and approximation coefficients at scale $s$ and $S$, respectively, obtained from the orthonormality properties of the wavelet and scaling functions. In this formulation,  $d=(1,2,3)$ is a wavelet directionality index, and $s=(1,2\dots,S)$ is a scale exponent, with $S=\log_2N$ the number of resolution levels allowed by the grid ($5$ for $32{\times}32$ images). Similarly, $\bxs=2^{s-1}(i\Delta,j\Delta)$ is a scale-dependent wavelet grid of $(N/2^s){\times}(N/2^s)$ elements where the basis functions are centered, with $i$, $j=1,3,\dots,N/2^{s-1}-1$. The wavelet coefficients represent the local fluctuations of $I$ centered at $\bxs$ at scale $s$, while the approximation coefficient is proportional to the global mean of $I$. At each scale, the filter width of the wavelets is $2^s \Delta$.

In this study, the 2D orthonormal basis functions $\psi^{(s,d)}(\bx-\bxs)$ are products of one-dimensional (1D) Haar wavelets \cite{Meneveau91}. The definition of 2D wavelets as multiplicative products of 1D wavelets is a particular choice that follows the MRA formulation \cite{Mallat89}. Haar wavelets have a narrow spatial support that provides a high degree of spatial localization. However, they display large spectral leakage at high wavenumbers since infinite spectral and spatial resolutions cannot be simultaneously attained due to limitations imposed by the uncertainty principle \cite{Adison}. Different boundary conditions can be assumed for the field $I$. We do not require such a choice in this study as we restrict ourselves to square images. However, the wavelet MRA framework is not limited to square inputs and can be generalized to rectangular inputs \cite{Adison}. 

The definition of 2D wavelets as multiplicative products of 1D wavelets is a particular choice that follows the MRA formulation described by Mallat \cite{Mallat89}, in which, the multivariate wavelets are characterized by an isotropic scale and therefore render limited information about anisotropy in the image. A large number of alternative basis functions have been recently proposed for replacing traditional wavelets when analyzing multi-dimensional data that exhibit complex anisotropic structures such as filaments and sheets. These include, but are not limited to, curvelets, contourlets, and shearlets \cite{kutyniok2012shearlets}. 

\subsection{Step 2: Image Filtering in Wavelet Space}
\label{prop_sp_FILT}

The second step decomposes $I$ as 
\begin{equation}
I = I_> + I_\leq, \label{decomp}
\end{equation}
where the filtered $I_>$ and remainder $I_\leq$ components correspond to the highest and lowest energetic wavelet modes of $I$, respectively. By construction, these two components are not spatially cross-correlated, as implied by the orthogonality of the wavelets and by the filtering operation described below. Note that large wavelet coefficients are associated with large fluctuations within the corresponding region of the scale-dependent wavelet grid $\boldsymbol{x_s}$, these being markers of underlying coherent structures. Under the assumptions that $I_\leq$ is additive Gaussian white noise, Donoho and Johnstone described a wavelet-based algorithm that is optimal for achieving the target decomposition (\ref{decomp}), since it minimizes the maximum $\mathbb{L}^2$-estimation error of $I_>$ \cite{Donoho94}. $I_>$ is obtained by retaining only the wavelet coefficients $\widecheck{I}^{(s,d)}$ whose absolute values satisfy
\begin{equation}
\widecheck{I}_>^{(s,d)}(\xsb) =
\left\{
	\begin{array}{ll}
		\widecheck{I}^{(s,d)}(\xsb)  & \mbox{if } |\widecheck{I}^{(s,d)}(\xsb)| \geq T, \\
		0 & \mbox{otherwise,} 
	\end{array}
\right. \label{filtering}
\end{equation}
for all scales $s$, positions $\boldsymbol{x_s}$ and directions $d$. In (\ref{filtering}), $T$ is a theoretical threshold defined as
\begin{equation}
T = \sqrt{2 \sigma^2_{I_\leq} \ln{N^2}}, \label{thresh}
\end{equation}
where $\sigma_{I_\leq}^2$ is the unknown variance of $I_\leq$. In this study, the iterative method of Azzalini et al. is employed, which converges to $T$ starting from a first iteration where $\sigma_{I_\leq}^2$ in (\ref{thresh}) is substituted by the variance $\sigma_{I}^2$ of the total image $I$ \cite{azzalini2005nonlinear}. This iterative procedure does not introduce significant computational overhead, since only one wavelet transform is required independently of the number of iterations. The algorithm does not introduce any hyperparameter when the theoretical threshold value is used. Note that the threshold is image-dependent, thereby ensuring that the algorithm adapts the number of superpixels to each image appropriately. The above filtering operation is equivalent to applying a binary filter mask to wavelet coefficients, denoted as wavelet-filter mask below.

The iterative method is deemed as converged when the relative variation in the estimated threshold $T$ is less than $0.1\%$ across consecutive iterations. A maximum of $\mathcal{O}(10)$ iterations were required to obtain the results presented in this paper. The overall computational cost is $\mathcal{O}(n_iM)$, where $n_i$ is the number of iterations and $M$ is the number of pixels in the image \cite{azzalini2005nonlinear}. In this work, we allow for further reduction in number of superpixels by varying the threshold $T$ to take larger values. \autoref{fig_framework} illustrates the application of this wavelet filtering method on an RGB image (filtering is applied to each channel independently) \cite{everingham2010pascal}. Most of the structural and edge information is preserved at all scales. However, a drawback of the method is that the superpixel boundaries are necessarily regular and axis-aligned.

\subsection{Step 3: Generating Superpixel Mesh from Wavelet-filter Mask}\label{prop_sp_GRP}

\begin{figure}[!htb]
    \centering	
    \includegraphics[clip,angle=90,width=0.9\textwidth]{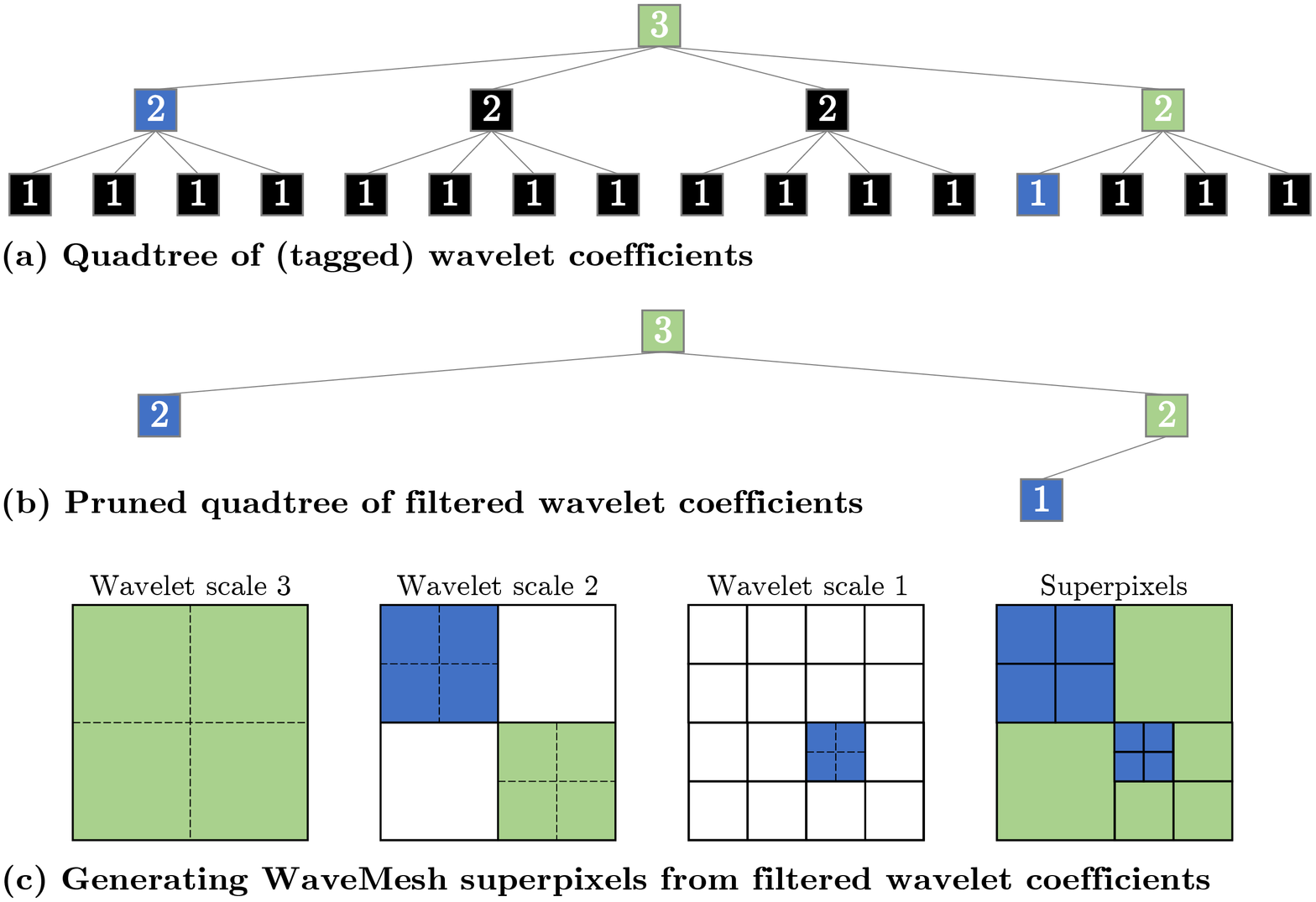}
    \caption{Illustration of the wavelet-based quadtree compression algorithm for an $8{\times}8$ image, along with the resulting adapted grid. Starting from the coarsest possible wavelet grid that contains just one superpixel, the algorithm adapts the grid by recursively splitting it. If the wavelet coefficient corresponding to a region is tagged (shown in blue), then that region is split into $2{\times}2$ superpixels.}
    \label{fig_quadtree}
\end{figure}

To generate superpixels for a given image, the final step is a grid adaptation based on the wavelet-filter mask described in \autoref{prop_sp_FILT}. The result is a non-uniform grid of multiscale superpixels adapted around regions of the image with high variability.

\textbf{Quadtree representation.} The algorithm is perhaps best understood by representing the wavelet coefficients in a quadtree \cite{finkel1974quad}, a tree data structure in which each node has exactly four children. A quadtree-based representation of wavelet coefficients was previously shown to be an efficient data structure for wavelet-based image compression \cite{banham1992wavelet,wakin2003geometric}. Here, the height of this quadtree equals the number of decomposition levels $S$ in the wavelet transform. Each vertex at a given level $s$ is associated with a triplet of wavelet coefficients $\textstyle [\widecheck{I}^{(s,d=0)}(\xsb),\widecheck{I}^{(s,d=1)}(\xsb),\widecheck{I}^{(s,d=2)}(\xsb)]$. All vertices from a given level correspond to wavelet coefficients across all locations at a given scale. The children vertices of a root vertex are the wavelet coefficients from that region in space at smaller scales. The quadtree representation of the wavelet coefficients of an $8{\times}8$ image is schematically represented in \autoref{fig_quadtree}(a). The number on each vertex indicates the scale, from the smallest scale $s{=}1$ associated with $2{\times}2$ pixel patches up to the largest scale $s{=}3$ associated with the entire $8{\times}8$ image. The pixel regions associated with each wavelet coefficient are delineated by solid lines in the three leftmost figures in \autoref{fig_quadtree}(c).

\textbf{Node tagging.} The vertices in the tree are tagged according to the filtering algorithm described in \autoref{prop_sp_FILT}. The tagged elements of the tree denoted by blue filled color in \autoref{fig_quadtree}(a,b) correspond to those with absolute values larger than the threshold $T$, and therefore correspond to locations in the image with important spatial variability. In the 2D case, tagging is applied if at least one of the 3 wavelet coefficients of $I$ per location is larger than the threshold. Additional tagging by green-filled color is applied to wavelet coefficients that are smaller than the threshold $T$ but that correspond to a spatial region with at least one tagged wavelet coefficient at a smaller scale. This corresponds to tagging all the ancestors of previously tagged vertices. This tagging procedure enforces cubic superpixels by ensuring that when there is a coherent structure at scale $s$ but not at a larger scale $s{+}1$, the wavelet coefficient at scale $s{+}1$ at that location are also tagged, hence triggering local grid refinement at level $s{+}1$. Non-tagged vertices are pruned as shown in \autoref{fig_quadtree}(b).

\textbf{Mesh generation.} Starting from the coarsest possible wavelet grid $\boldsymbol{x_s}=\boldsymbol{x_S}$ that contains just one superpixel, the algorithm adapts the grid by recursively splitting it as follows. If the wavelet coefficient corresponding to a region is tagged, then that region is split into $2{\times}2$ superpixels, which locally refines the grid. The algorithm is stopped otherwise. The same recursive loop is then applied to the refined superpixels. The final configuration of the adapted grid is obtained when none of the wavelet coefficients in any the superpixels are tagged. An example of final adapted grid is shown in \autoref{fig_quadtree}(c). The dashed lines correspond to the superpixel refinement due to the vertex being tagged. Adapted grid from a natural image is shown in \autoref{fig_framework} where the superpixel mesh exhibits desired level of heterogeneity with  multiscale refinement around edges. For RGB images, the most restrictive mesh is employed at every location and scale. In other words, tagging for the full image is applied if at least of the channels is tagged. 

\section{WavePool: Spatially heterogeneous pooling }
\label{prop_GNN_pool}

\begin{figure}[!htb]
    \centering	
    \includegraphics[clip,angle=-90,width=0.8\textwidth]{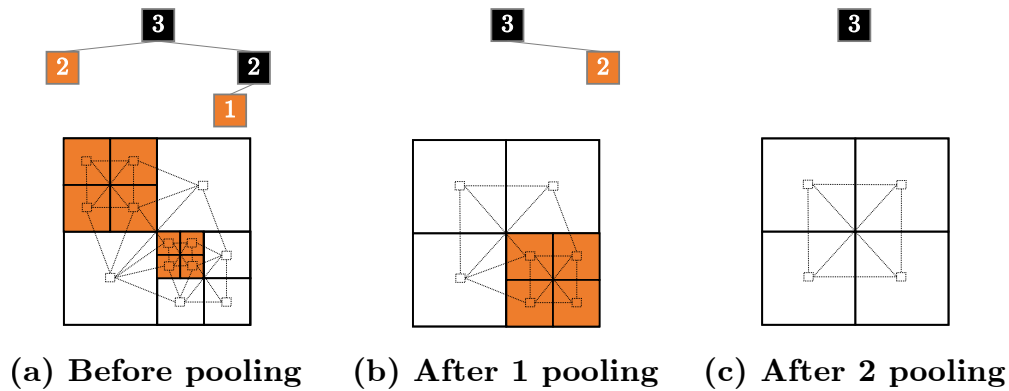}
    \caption{Illustration of WavePool from wavelet quadtree representation. Leaf nodes ($2{\times}2$ superpixels)  are recursively pooled. In the lower panel, dashed squares and lines correspond to nodes and edges in the superpixel graph.}\label{fig_pooling}
\end{figure}

The proposed spatially heterogeneous pooling, WavePool, is best explained using the wavelet coefficient quadtree representation described in \autoref{prop_sp_GRP}. One WavePool operation involves aggregating all the leaf nodes of the wavelet quadtree. In the pixel domain, this step corresponds to merging patches of $2{\times}2$ superpixels into a parent superpixel, and aggregating the node features with a choice of pooling function (e.g. max). \autoref{fig_pooling} illustrates WavePool on a simple superpixel mesh and shows its effect on both the quadtree (upper panel) and region adjacency graph (lower panel) representation. In a region adjacency graph (RAG), nodes represent superpixel centroids, and edges connect neighboring superpixels. Note that GNNs are trained on RAGs. RAG is not a tree and should not be confused with the wavelet coefficient quadtree. 

\begin{figure}[!htb]
    \centering	
    \includegraphics[clip,angle=90,width=0.6\textwidth]{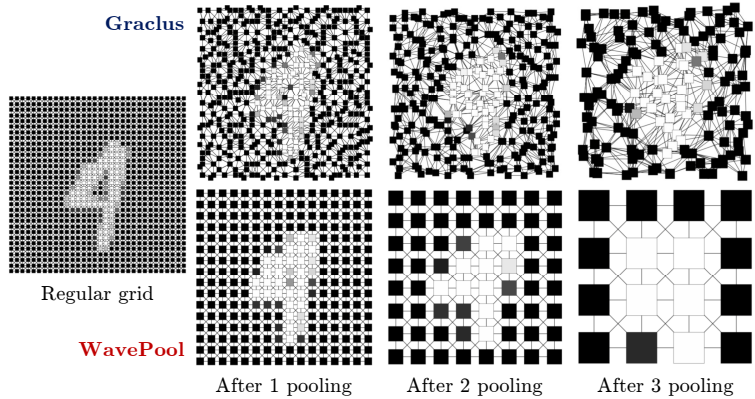}
    \caption{WavePool vs. GraclusPool on a regular grid.}
    \label{fig:pooling_property}
\end{figure}

WavePool generalizes the classical CNN pooling operation. For a regular-pixel grid as in \autoref{fig:pooling_property}, WavePool exactly matches the $2{\times}2$ pooling in CNN. However, this is not true with GraclusPool. Although more general than its CNN counterpart, WavePool is restricted to WaveMesh or more broadly to any quadtree based superpixels \cite{tanimoto1975hierarchical, zhang2018image}, unlike GraclusPool. 

\begin{figure}[!htb]
\centering
\includegraphics[scale=0.3]{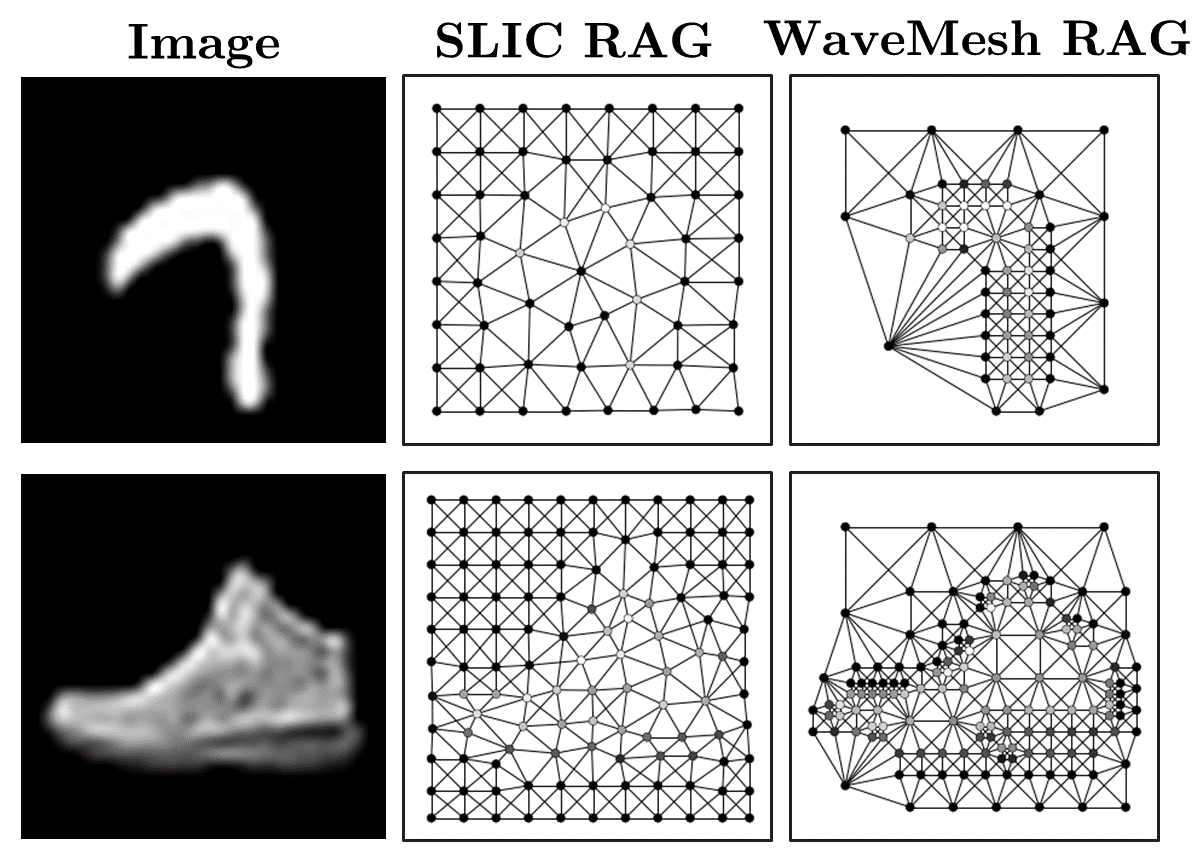}
\caption{MNIST and Fashion-MNIST images with SLIC and WaveMesh RAG. Nodes represent superpixel centroids. SLIC and WaveMesh graphs are structurally different. In WaveMesh, there are more nodes along the object boundaries and fewer nodes in regions without much variation.}
\label{fig:all_datasets_img_graph}
\end{figure}

\section{Experimental setup}
\label{sec:experiments}

\subsection{Datasets}
\label{subsec:datasets}
To compare WaveMesh and similar-sized superpixels we perform experiments on three datasets: MNIST, Fashion-MNIST, and CIFAR-10 \cite{lecun1998gradient,xiao2017fashion,krizhevsky2009learning}. We represent superpixels by RAGs as shown in \autoref{fig:all_datasets_img_graph}, where mean intensity of superpixel is used as a node feature. Edges in the graph are directed with pseudo-coordinates as in \cite{fey2018splinecnn}. 

\subsection{Model: SplineCNN}
\label{subsec:model_splinecnn}
\textbf{Why SplineCNN?} We use SplineCNN \cite{fey2018splinecnn}, in all our experiments. SplineCNN is an ideal candidate for this study for the following reasons. First, it is a state-of-the-art GNN for image graph classification. Second, Fey et al. report that edge detecting patterns are learned by the kernels in SplineCNN when trained on superpixels. Third, the spline convolution (SConv) operator is a generalization of the convolution operator in CNNs with odd kernel size. This property of SConv operator combines nicely with WavePool that naturally collapses to classical CNN pooling on a regular pixel-grid.

\textbf{SplineCNN configurations.} We conduct experiments on two SplineCNN configurations using the implementation available in PyTorch Geometric \cite{Fey/Lenssen/2019}. The configurations are: 
\begin{enumerate}[itemsep=-2pt,topsep=0pt,leftmargin=*]
\item SConv$((3,3),1,32)$ $\rightarrow$ Pool $\rightarrow$ SConv$((3,3),32,64)$ $\rightarrow$  Pool $\rightarrow$ Global mean pool $\rightarrow$ FC$(128)$ $\rightarrow$ FC$(10)$. Network has 30506 parameters. 
\item SConv$((3,3),1,32)$ $\rightarrow$ Pool $\rightarrow$ SConv$((3,3),32,64)$ $\rightarrow$ Pool $\rightarrow$ SConv$((3,3),64,128)$ $\rightarrow$ Pool $\rightarrow$ Global mean pool $\rightarrow$ FC$(256)$ $\rightarrow$ FC$(10)$. Network has 139178 parameters. 
\end{enumerate}
Here, FC denotes a fully-connected layer and Pool denotes a GraclusPool or WavePool layer or no pooling.

\subsection{Comparison with SLIC}
\label{subsec:comp_slic}
The SLIC superpixeling algorithm is based on $k$-means clustering \cite{achanta2012slic}. We compare the performance of WaveMesh with SLIC. Many superpixeling algorithms have been proposed since SLIC, however it is still an ideal baseline for this study for the following reasons.
\begin{itemize}[itemsep=-2pt,topsep=0pt,leftmargin=*]
\item SLIC is one among the six algorithms recommended by Stutz et al.\ after evaluating 28 state-of-the-art superpixeling algorithms \cite{stutz2018superpixels}. All the recommended algorithms show superior performance in Boundary Recall, Undersegmentation Error, and Explained Variation (EV). 
\item Giraud et al.\ evaluates Achievable Segmentation Accuracy (ASA) of nine state-of-the-art superpixeling algorithms. SLIC is among the top six in ASA \cite{giraud2019evaluation}. Also, the difference in ASA among top six algorithms is small.
\item SLIC superpixels are used in all prior GNN studies \cite{monti2017geometric, fey2018splinecnn, knyazev2019image, knyazev2019understanding, dwivedi2020benchmarking, avelar2020superpixel, mesquita2020rethinking}. Moreover, our goal is to compare multiscale WaveMesh superpixels with routinely used similar-sized superpixels for image classification, and not to find the best superpixeling algorithm for image classification using GNNs.  
\end{itemize}

\begin{table}[!htb]
\caption{Results on MNIST superpixels. For each group of experiments, \red{lowest value} is marked in red. Experiments R1 and R2 are from \cite{monti2017geometric} and \cite{fey2018splinecnn}. Experiments R3--R5 from \cite{dwivedi2020benchmarking} are for models RingGNN, MoNet and GatedGCN. They report min and max value in \textbf{\#Nodes}, and number of parameters in \textbf{Config}.}
\small
\centering
\begin{tabular}{rlccccc}
\toprule 
\textbf{\#} &  \textbf{SP} & \textbf{\#Nodes} & \textbf{Config} & \textbf{Pool} & \textbf{Train acc (\%)} & \textbf{Test acc (\%)}  \\
\midrule
1 & WM & 238$\pm$50 & 1 & NP & 92.30$\pm$0.43 & 92.60$\pm$0.45 \\

2 & WM  & 238$\pm$50 & 1 & GR & 92.33$\pm$0.09 & \red{89.63$\pm$0.45} \\ 

3 & WM & 238$\pm$50 & 1 & WP & 95.75$\pm$0.08 & 95.44$\pm$0.12
\\ 
\midrule
4 & WM & 238$\pm$50 & 2 & NP & 98.23$\pm$0.07 & 97.70$\pm$0.10 \\ 
5 & WM  & 238$\pm$50  & 2 & GR & 98.39$\pm$ 0.05 & \red{96.80$\pm$0.11} \\ 

6 & WM & 238$\pm$50 & 2 & WP & 99.68$\pm$ 0.03 & 98.68$\pm$0.08 \\
\midrule 
7 & SL & 241$\pm$5 & 1 & NP & 95.85$\pm$0.03 & 95.99$\pm$0.07 \\ 
8 & SL  & 241$\pm$5  & 1 & GR & 95.50$\pm$0.21 & 95.51$\pm$0.29 	\\ 

\midrule 
9 & SL & 241$\pm$5 & 2 & NP & 99.56$\pm$0.03 & 98.79$\pm$0.07 \\
10 & SL & 241$\pm$5 & 2 & GR & 98.07$\pm$0.04 & \red{97.83$\pm$0.11}  \\
\midrule

11 & WM & 57$\pm$12 & 1 & NP & 95.66$\pm$0.06  & 95.54$\pm$0.15 \\ 
12 & WM  & 57$\pm$12  & 1 & GR & 93.34$\pm$0.04 & \red{92.53$\pm$0.15} 
\\ 
13 & WM & 57$\pm$12 & 1 & WP & 96.30$\pm$0.10 & 93.74$\pm$0.17  \\ 

\midrule
14 & WM & 57$\pm$12 & 2 & NP & 98.74$\pm$0.06 & 97.53$\pm$0.09\\ 
15 & WM  & 57$\pm$12  & 2 & GR & 95.68$\pm$0.09 & \red{94.21$\pm$0.21} \\ 

16 & WM & 57$\pm$12 & 2 & WP & 99.23$\pm$0.04 & \red{93.84$\pm$0.48} \\

\midrule 
17 & SL & 59$\pm$2 & 1 & NP & 95.56$\pm$0.11 & 95.17$\pm$0.12\\
18 & SL  & 59$\pm$2  & 1 & GR & 92.34$\pm$0.11 & \red{91.18$\pm$0.22}  \\ 

\midrule
19 & SL & 59$\pm$2 & 2 & NP & 98.84$\pm$0.06 & 97.18$\pm$0.07 \\
20 & SL & 59$\pm$2 & 2 & GR & 94.13$\pm$0.08 & \red{92.99$\pm$0.22} 	\\
\midrule 
R1 & SL & 75$\pm$0 & -- & GR & -- & 91.11 \\

R2 & SL & 75$\pm$0 & -- & GR & -- & 95.22  \\

R3 & SL & 40--75 & 105398 & -- & 11.24$\pm$0.00 & \red{11.35$\pm$0.00}  \\

R4 & SL & 40--75 & 104049 & -- & 96.61$\pm$0.44 & 90.81$\pm$0.03\\ 

R5 & SL & 40--75 & 104217 & -- & 100.00$\pm$0.00 & 97.34$\pm$0.14\\ 
\bottomrule
\end{tabular}
\label{tab:mnist_results}
\end{table}

\subsection{Evaluation}
The two main objectives of our experiments are as follows. First, to understand the performance of WaveMesh superpixels under three different local-pooling settings, everything else being the same. Second, to understand how SplineCNN performs on SLIC and WaveMesh superpixels under the same network architecture and training settings. Together, these objectives systematically explore image classification using multiscale superpixels (WaveMesh) compared to single-scale superpixels (SLIC). Using the two SplineCNN configurations mentioned in \autoref{subsec:model_splinecnn} we perform extensive experiments on WaveMesh and SLIC superpixels by varying the following. First, the number of superpixels. Second, changing the local-pooling: no pooling, GraclusPool, and WavePool.

The SplineCNN implementation in PyTorch Geometric uses Adam optimizer with an initial learning rate of 0.01, which is decreased by a factor of 10 after 15 and 25 epochs. Since the goal of our experiments is not to tune the best model for WaveMesh superpixels, we use the default hyperparameters from their implementation. We train the network for 30 epochs on MNIST and Fashion-MNIST and 75 epochs on CIFAR-10. The pooling function is max for both WavePool and GraclusPool.  All experiments are repeated five times. 

We also compare WaveMesh and SLIC superpixels on two traditional superpixel evaluation metrics: ASA and EV. While there are many superpixel evaluation metrics, we use ASA and EV because Giruad et al.\ recommends ASA to evaluate adherence to object boundaries, and EV to evaluate the color homogeneity within superpixels \cite{giraud2019evaluation}. We calculate ASA and EV on the BSD300 dataset for SLIC and WaveMesh superpixels using the code provided by \cite{giraud2019evaluation}. Default compactness value of 10 is used in the SLIC algorithm, and approximately 500 superpixels are generated for each image in the dataset \cite{martin2001databaseBSD300}. 

\begin{table}
\caption{Results on Fashion-MNIST superpixels. For each group of experiments, \red{lowest value} is marked in red. Experiment R1 is from \cite{avelar2020superpixel}.}
\small
\centering
\begin{tabular}{rlcccccc}
\toprule 
\textbf{\#} &  \textbf{SP} & \textbf{\#Nodes} & \textbf{Config} & \textbf{Pool} & \textbf{Train acc (\%)} & \textbf{Test acc (\%)} \\
\midrule
1 & WM & 436$\pm$129 & 1  & NP & 80.34$\pm$0.42 & 79.60$\pm$0.44  \\ 
2 & WM  & 436$\pm$129 & 1 & GR & 80.36$\pm$0.39 & \red{65.35$\pm$2.94}  \\ 

3 & WM & 436$\pm$129 & 1 & WP & 85.77$\pm$0.18 & 76.60$\pm$0.83  \\ 

\midrule
4 & WM & 436$\pm$129 & 2 & NP & 86.86$\pm$0.15 & 85.71$\pm$0.16 \\ 
5 & WM  & 436$\pm$129 & 2 & GR & 85.40$\pm$0.10 & \red{75.69$\pm$1.47} \\ 

6 & WM & 436$\pm$129 & 2 & WP & 92.58$\pm$0.07 & 83.66$\pm$1.49  \\ 

\midrule

7 & WM & 261$\pm$35 & 1 & NP & 82.54$\pm$0.20 & 81.61$\pm$0.24  \\ 
8 & WM  & 261$\pm$35  & 1 & GR & 81.32$\pm$0.13 & \red{76.75$\pm$ 0.33}  \\ 

9 & WM & 261$\pm$35 & 1 & WP & 85.91$\pm$0.10 & 81.35$\pm$0.69  \\ 

\midrule
10 & WM & 261$\pm$35 & 2 & NP &  88.20$\pm$0.24 & 86.60$\pm$0.14 \\ 
11 & WM  & 261$\pm$35  & 2 & GR & 85.18$\pm$0.18 & \red{79.78$\pm$0.46}  \\ 

12 & WM & 261$\pm$35 & 2 & WP & 92.34$\pm$0.15 & 87.65$\pm$0.36  \\

\midrule 
13 & SL & 259$\pm$7 & 1 & NP & 83.60$\pm$0.16 & 82.37$\pm$0.25 \\ 
14 & SL  & 259$\pm$7  & 1 & GR & 82.91$\pm$0.08 & 81.49$\pm$0.38 \\ 

\midrule 
15 & SL & 259$\pm$7 & 2 & NP & 89.01$\pm$0.31 & 87.29$\pm$0.30 \\ 
16 & SL & 259$\pm$7 & 2 & GR & 86.71$\pm$0.10 & \red{85.00$\pm$0.32}  \\

\midrule
17 & WM & 134$\pm$22 & 1 & NP  & 83.23$\pm$0.09 & 82.04$\pm$0.10\\ 
18 & WM  & 134$\pm$22  & 1 & GR & 80.92$\pm$0.16 & \red{78.85$\pm$0.09} \\ 

19 & WM & 134$\pm$22 & 1 & WP & 85.18$\pm$0.13 & 80.42$\pm$0.33 \\ 

\midrule
20 & WM & 134$\pm$22 & 2 & NP & 87.88$\pm$0.07 & 85.62$\pm$0.26 \\ 
21 & WM  & 134$\pm$22  & 2 & GR & 83.84$\pm$0.13 & \red{80.60$\pm$0.21} \\ 

22 & WM & 134$\pm$22 & 2 & WP & 90.98$\pm$0.13 & 82.65$\pm$0.52\\

\midrule 
23 & SL & 118$\pm$4  & 1  & NP & 83.01$\pm$0.15 & 81.44$\pm$0.19 \\ 
24 & SL  & 118$\pm$4  & 1 & GR & 81.46$\pm$0.16 & \red{79.59$\pm$0.34} \\ 

\midrule 
25 & SL & 118$\pm$4  & 2 & NP & 88.31$\pm$0.19 & 86.10$\pm$0.25 \\ 
26 & SL &  118$\pm$4 & 2 & GR & 84.11$\pm$0.15 & \red{82.12$\pm$0.27}  \\
\midrule 
R1 & SL  & $\leq 75$ & -- & -- & -- & 83.07 \\
\bottomrule
\end{tabular}
\label{tab:fashionmnist_test_acc}
\end{table}

\section{Results and Discussion}

\textbf{Format of Tables 1--3.} Classification results on MNIST, Fashion-MNIST, and CIFAR-10 graphs are reported in Tables 1--3. Experiment numbers starting with `R' report results from prior studies. For brevity, we use the acronyms SP: superpixel, WM: WaveMesh, SL: SLIC, NP: no pooling, GR: GraclusPool, and WP: WavePool. In each table, experiments are partitioned into groups (separated by a mid-rule). All experiments within a group are identical except for pooling. For each experiment, we report the mean and standard deviation values for number of nodes (superpixels), train and test accuracy. \red{Lowest} test accuracy within a group is highlighted in red.

\textbf{Number of superpixels.} Experiments 1--6 in Tables 1--3 uses WaveMesh superpixels obtained using the theoretical threshold $T$. The other WaveMesh experiments are on fewer superpixels obtained by scaling (increasing) $T$. To perform one-to-one comparison we generate approximately the same number of superpixels using the SLIC implementation in scikit-learn. We were unable to generate greater than $\approx$250 superpixels for FashionMNIST using SLIC. 

\textbf{Effect of pooling.} In all WaveMesh experiments test accuracy is the lowest while using GraclusPool. Comparing WaveMesh+WavePool with WaveMesh+Graclus, the former performs significantly better in majority of the experiments. Therefore, cluster assignment has an effect on the performance of SplineCNN while using multiscale WaveMesh superpixels. This is unlike what was observed in \cite{mesquita2020rethinking} with MNIST single-scale SLIC superpixels on other GNN models. When comparing WaveMesh+NoPooling with WaveMesh+WavePool there is no clear winner. Similarly, SLIC+NoPooling
is atleast as good as SLIC+Graclus. Therefore, while training a model on superpixels it is good to begin with a model that has just convolution layers. 

\textbf{WaveMesh versus SLIC.} When comparing test accuracy in the absence of pooling WaveMesh is better than SLIC for CIFAR-10, better or same as SLIC for Fashion-MNIST, and there is no clear trend for MNIST. In the presence of pooling, WaveMesh+WavePool is better than or on-par with SLIC+Graclus. Recalling the objectives of this study, we conclude that the performance of SplineCNN with multiscale WaveMesh superpixels is just as good as SLIC superpixels.

\textbf{Performance with other GNNs}. Dwivedi et al.\ benchmark the performance of GNN models on MNIST and CIFAR-10 SLIC superpixels \cite{dwivedi2020benchmarking}.  From their results, RingGNN and GatedGCN perform the worst and best, respectively. Therefore, it is clear that not all GNN models perform well on SLIC superpixels. While we have not trained these GNN models with WaveMesh superpixels, we expect a similar trend with WaveMesh. Results from their paper for RingGNN, MoNet and GatedGCN are reported in \autoref{tab:mnist_results} (experiment R3--R5) and \autoref{tab:cifar10_test_acc} (experiment R1--R3). Results for MoNet are shown because SplineCNN builds on the work of MoNet.

\textbf{ASA and EV.} \autoref{tab:ev_asa} shows ASA and EV averaged across images in the BSD300 dataset. Giruad et al.\ recommends ASA to evaluate adherence to object boundaries, and EV to evaluate the color homogeneity within superpixels \cite{giraud2019evaluation}. WaveMesh is comparable with SLIC on ASA and inferior on EV.

\begin{table}[!htb]
\caption{Results on CIFAR-10 superpixels. For each group of experiments, \red{lowest value} is marked in red. Experiments R1--R3 from \cite{dwivedi2020benchmarking} are for models RingGNN, MoNet and GatedGCN. They report min and max value in \textbf{\#Nodes}, and number of parameters in \textbf{Config}.}

\small
\centering
\begin{tabular}{rlccccc}
\toprule 
\textbf{\#} &  \textbf{SP} & \textbf{\#Nodes} & \textbf{Config} & \textbf{Pool} & \textbf{Train acc (\%)} & \textbf{Test acc (\%)} \\
\midrule
1 & WM  & 197$\pm$82 & 1 & NP & 51.18$\pm$0.15  & 50.59$\pm$0.17  \\ 

2 & WM  & 197$\pm$82 & 1 & GR & 52.63$\pm$0.36  & \red{43.36$\pm$0.72}  \\ 

3 & WM & 197$\pm$82 & 1 & WP & 55.04$\pm$0.21 & 52.58$\pm$0.21 \\ 

\midrule
4 & WM  & 197$\pm$82  & 2 & NP & 61.52$\pm$0.36  & 58.33$\pm$0.40  \\
5 & WM  & 197$\pm$82  & 2 & GR & 60.28$\pm$0.18  & \red{50.42$\pm$0.27} \\ 

6 & WM & 197$\pm$82 & 2 & WP & 70.25$\pm$0.30 & 56.89$\pm$0.31  \\ 

\midrule
7 & SL  & 215$\pm$15  & 1 &  NP & 48.37$\pm$0.24   & 47.25$\pm$0.21 \\ 
8 & SL  & 215$\pm$15  & 1 & GR &  50.96$\pm$ 0.51  & \red{45.87$\pm$0.28}  \\ 
\midrule 

9 & SL & 215$\pm$15 & 2 & NP & 58.61$\pm$0.36   & 56.60$\pm$0.18 \\
10 & SL & 215$\pm$15 & 2 & GR & 59.09$\pm$ 0.20  & \red{50.69$\pm$0.45}  \\
\midrule 
R1 & SL & 85--150 & 105165 & -- & 19.56$\pm$16.40 & \red{19.30$\pm$16.12}  \\

R2 & SL & 85--150 & 104229 & -- & 65.92$\pm$2.52 & 54.66$\pm$0.52\\ 

R3 & SL & 85--150 & 104357 & -- & 94.55$\pm$1.02 & 67.31$\pm$0.31\\ 
\bottomrule
\end{tabular}
\label{tab:cifar10_test_acc}
\end{table}

\begin{table}[!htb]
\small
\centering
\caption{Performance on BSD300 dataset.}
\begin{tabular}{lcc}
\toprule
\textbf{Metric} & SLIC  & WaveMesh \\ \midrule 
Num superpixels & 475$\pm$20 & 515$\pm$129 \\
Achievable Segmentation Accuracy (ASA) & 0.967$\pm$0.014 & 0.950$\pm$0.025\\
Explained Variation (EV) & 0.870$\pm$0.093 & 0.783$\pm$0.128 \\ 
 \bottomrule
\end{tabular}
\label{tab:ev_asa}
\end{table}

\section{Conclusion}\label{conclusions}

Prior GNN studies on image graph classification have been restricted to graphs that represent a regular grid or similar-sized SLIC superpixels. 
To fill this gap, we investigated image classification using multiscale superpixels and SplineCNN. We proposed 1) WaveMesh, a novel wavelet-based superpixeling algorithm, where the number and sizes of superpixels in an image are computed based on its content, and 2) WavePool, a novel spatially heterogeneous pooling scheme tailored to WaveMesh superpixels. Due to the multiscale nature of WaveMesh superpixels, their RAGs are structurally different from those of SLIC. Extensive experiments on benchmark datasets show that poor choice of local-pooling negatively affects the performance of SplineCNN while using WaveMesh superpixels. We also show that SplineCNN learns from multiscale WaveMesh superpixels on-par with SLIC superpixels under the same setting. Further investigation similar to \cite{stutz2018superpixels} and \cite{giraud2019evaluation} is required to rank superpixeling algorithms for image graph classification using popular GNNs.

\clearpage

\printbibliography

@inproceedings{Fey/Lenssen/2019,
  title={Fast Graph Representation Learning with {PyTorch Geometric}},
  author={Fey, Matthias and Lenssen, Jan E.},
  booktitle={ICLR Workshop on Representation Learning on Graphs and Manifolds},
  year={2019},
}

@article{donoho1994ideal,
  title={Ideal spatial adaptation by wavelet shrinkage},
  author={Donoho, David L and Johnstone, Iain M},
  journal={biometrika},
  volume={81},
  number={3},
  pages={425--455},
  year={1994},
  publisher={Oxford University Press}
}

@article{azzalini2005nonlinear,
  title={Nonlinear wavelet thresholding: A recursive method to determine the optimal denoising threshold},
  author={Azzalini, Alexandre and Farge, Marie and Schneider, Kai},
  journal={Applied and Computational Harmonic Analysis},
  volume={18},
  number={2},
  pages={177--185},
  year={2005},
  publisher={Elsevier}
}

@inproceedings{fey2018splinecnn,
  title={SplineCNN: Fast geometric deep learning with continuous B-spline kernels},
  author={Fey, Matthias and Eric Lenssen, Jan and Weichert, Frank and M{\"u}ller, Heinrich},
  booktitle={Proceedings of the IEEE Conference on Computer Vision and Pattern Recognition},
  pages={869--877},
  year={2018}
}

@article{sabottke2020effect,
  title={The effect of image resolution on deep learning in radiography},
  author={Sabottke, Carl F and Spieler, Bradley M},
  journal={Radiology: Artificial Intelligence},
  volume={2},
  number={1},
  pages={e190015},
  year={2020},
  publisher={Radiological Society of North America}
}

@article{lakhani2020importance,
  title={The Importance of Image Resolution in Building Deep Learning Models for Medical Imaging},
  author={Lakhani, Paras},
  journal={Radiology: Artificial Intelligence},
  volume={2},
  number={1},
  pages={e190177},
  year={2020},
  publisher={Radiological Society of North America}
}

@book{Adison,
  title={The illustrated wavelet transform handbook: introductory theory and applications in science, engineering, medicine and finance},
  author={Addison, Paul S},
  year={2017},
  publisher={CRC press}
}

@article{Mallat89,
  title={A theory for multiresolution signal decomposition: the wavelet representation},
  author={Mallat, Stephane G},
  journal={IEEE T. Pattern Anal.},
  volume={11},
  number={7},
  pages={674--693},
  year={1989},
  publisher={Ieee}
}

@article{Meneveau91,
  title={Analysis of turbulence in the orthonormal wavelet representation},
  author={Meneveau, Charles},
  journal={J. Fluid Mech.},
  volume={232},
  pages={469--520},
  year={1991},
  publisher={Cambridge University Press}
}

@book{kutyniok2012shearlets,
  title={Shearlets: Multiscale analysis for multivariate data},
  author={Kutyniok, Gitta and Labate, Demetrio},
  year={2012},
  publisher={Springer Science \& Business Media}
}

@article{Donoho94,
    author = {Donoho, David L and Johnstone, Iain M},
    title = {Ideal spatial adaptation by wavelet shrinkage},
    journal = {Biometrika},
    year = {1994},
    volume = {81},
    pages = {425--455}
}

@article{Bassenne18wave,
  title={Wavelet multiresolution analysis of particle-laden turbulence},
  author={Bassenne, Maxime and Moin, Parviz and Urzay, Javier},
  journal={Phys. Rev. Fluids},
  volume={3},
  number={8},
  pages={084304},
  year={2018},
  publisher={APS}
}

@article{Schneider10,
  title={Wavelet methods in computational fluid dynamics},
  author={Schneider, Kai and Vasilyev, Oleg V},
  journal={Annu. Rev. Fluid Mech.},
  volume={42},
  pages={473--503},
  year={2010},
  publisher={Annual Reviews}
}

@article{finkel1974quad,
  title={Quad trees a data structure for retrieval on composite keys},
  author={Finkel, Raphael A. and Bentley, Jon Louis},
  journal={Acta informatica},
  volume={4},
  number={1},
  pages={1--9},
  year={1974},
  publisher={Springer}
}

@inproceedings{monti2017geometric,
  title={Geometric deep learning on graphs and manifolds using mixture model cnns},
  author={Monti, Federico and Boscaini, Davide and Masci, Jonathan and Rodola, Emanuele and Svoboda, Jan and Bronstein, Michael M},
  booktitle={Proceedings of the IEEE Conference on Computer Vision and Pattern Recognition},
  pages={5115--5124},
  year={2017}
}

@article{achanta2012slic,
  title={SLIC superpixels compared to state-of-the-art superpixel methods},
  author={Achanta, Radhakrishna and Shaji, Appu and Smith, Kevin and Lucchi, Aurelien and Fua, Pascal and S{\"u}sstrunk, Sabine},
  journal={IEEE transactions on pattern analysis and machine intelligence},
  volume={34},
  number={11},
  pages={2274--2282},
  year={2012},
  publisher={IEEE}
}

@inproceedings{ren2003learning,
  title={Learning a classification model for segmentation},
  author={Ren, Xiaofeng and Malik, Jitendra},
  booktitle={Proceedings Ninth IEEE International Conference on Computer Vision},
  pages={10},
  year={2003},
  organization={IEEE}
}

@inproceedings{knyazev2019understanding,
  title={Understanding attention and generalization in graph neural networks},
  author={Knyazev, Boris and Taylor, Graham W and Amer, Mohamed},
  booktitle={Advances in Neural Information Processing Systems},
  pages={4202--4212},
  year={2019}
}

@article{knyazev2019image,
  title={Image classification with hierarchical multigraph networks},
  author={Knyazev, Boris and Lin, Xiao and Amer, Mohamed R and Taylor, Graham W},
  journal={arXiv preprint arXiv:1907.09000},
  year={2019}
}

@article{dwivedi2020benchmarking,
  title={Benchmarking graph neural networks},
  author={Dwivedi, Vijay Prakash and Joshi, Chaitanya K and Laurent, Thomas and Bengio, Yoshua and Bresson, Xavier},
  journal={arXiv preprint arXiv:2003.00982},
  year={2020}
}

@article{dhillon2007weighted,
  title={Weighted graph cuts without eigenvectors a multilevel approach},
  author={Dhillon, Inderjit S and Guan, Yuqiang and Kulis, Brian},
  journal={IEEE transactions on pattern analysis and machine intelligence},
  volume={29},
  number={11},
  pages={1944--1957},
  year={2007},
  publisher={IEEE}
}

@inproceedings{defferrard2016convolutional,
  title={Convolutional neural networks on graphs with fast localized spectral filtering},
  author={Defferrard, Micha{\"e}l and Bresson, Xavier and Vandergheynst, Pierre},
  booktitle={Advances in neural information processing systems},
  pages={3844--3852},
  year={2016}
}

@article{lecun1998gradient,
  title={Gradient-based learning applied to document recognition},
  author={LeCun, Yann and Bottou, L{\'e}on and Bengio, Yoshua and Haffner, Patrick},
  journal={Proceedings of the IEEE},
  volume={86},
  number={11},
  pages={2278--2324},
  year={1998},
  publisher={Ieee}
}

@article{krizhevsky2009learning,
  title={Learning multiple layers of features from tiny images},
  author={Krizhevsky, Alex and Hinton, Geoffrey},
  year={2009},
  journal={Technical Report},
  publisher={Citeseer}
}

@article{xiao2017fashion,
  title={Fashion-mnist: a novel image dataset for benchmarking machine learning algorithms},
  author={Xiao, Han and Rasul, Kashif and Vollgraf, Roland},
  journal={arXiv preprint arXiv:1708.07747},
  year={2017}
}

@article{avelar2020superpixel,
  title={Superpixel Image Classification with Graph Attention Networks},
  author={Avelar, Pedro HC and Tavares, Anderson R and da Silveira, Thiago LT and Jung, Cl{\'a}udio R and Lamb, Lu{\'\i}s C},
  journal={arXiv preprint arXiv:2002.05544},
  year={2020}
}

@article{everingham2010pascal,
  title={The pascal visual object classes (voc) challenge},
  author={Everingham, Mark and Van Gool, Luc and Williams, Christopher KI and Winn, John and Zisserman, Andrew},
  journal={International journal of computer vision},
  volume={88},
  number={2},
  pages={303--338},
  year={2010},
  publisher={Springer}
}

@article{tanimoto1975hierarchical,
  title={A hierarchical data structure for picture processing},
  author={Tanimoto, Steven and Pavlidis, Theo},
  journal={Computer graphics and image processing},
  volume={4},
  number={2},
  pages={104--119},
  year={1975},
  publisher={Elsevier}
}

@article{zhang2018image,
  title={Image Segmentation Based on Multiscale Fast Spectral Clustering},
  author={Zhang, Chongyang and Zhu, Guofeng and Chen, Minxin and Chen, Hong and Wu, Chenjian},
  journal={arXiv preprint arXiv:1812.04816},
  year={2018}
}

@inproceedings{banham1992wavelet,
  title={A wavelet transform image coding technique with a quadtree structure},
  author={Banham, Mark R and Sullivan, Barry J},
  booktitle={[Proceedings] ICASSP-92: 1992 IEEE International Conference on Acoustics, Speech, and Signal Processing},
  volume={4},
  pages={653--656},
  year={1992},
  organization={IEEE}
}

@inproceedings{wakin2003geometric,
  title={Geometric methods for wavelet-based image compression},
  author={Wakin, Michael B and Romberg, Justin K and Choi, Hyeokho and Baraniuk, Richard G},
  booktitle={Wavelets: Applications in Signal and Image Processing X},
  volume={5207},
  pages={507--520},
  year={2003},
  organization={International Society for Optics and Photonics}
}

@article{mesquita2020rethinking,
  title={Rethinking pooling in graph neural networks},
  author={Mesquita, Diego and Souza, Amauri H and Kaski, Samuel},
  journal={arXiv preprint arXiv:2010.11418},
  year={2020}
}

@article{stutz2018superpixels,
  title={Superpixels: An evaluation of the state-of-the-art},
  author={Stutz, David and Hermans, Alexander and Leibe, Bastian},
  journal={Computer Vision and Image Understanding},
  volume={166},
  pages={1--27},
  year={2018},
  publisher={Elsevier}
}

@article{giraud2019evaluation,
  title={Evaluation Framework of Superpixel Methods with a Global Regularity Measure},
  author={Giraud, R{\'e}mi and Ta, Vinh-Thong and Papadakis, Nicolas},
  journal={arXiv preprint arXiv:1903.07162},
  year={2019}
}

@inproceedings{martin2001databaseBSD300,
  title={A database of human segmented natural images and its application to evaluating segmentation algorithms and measuring ecological statistics},
  author={Martin, David and Fowlkes, Charless and Tal, Doron and Malik, Jitendra},
  booktitle={Proceedings Eighth IEEE International Conference on Computer Vision. ICCV 2001},
  volume={2},
  pages={416--423},
  year={2001},
  organization={IEEE}
}
\end{document}